\documentclass[
]{ceurart}

\usepackage{listings}
\usepackage{float}
\usepackage{graphicx}
\usepackage{wrapfig}
\usepackage{booktabs}
\usepackage{tabularx}

\usepackage{booktabs}
\usepackage{listings}
\usepackage{tabularx}
\usepackage[most]{tcolorbox}

\newtcolorbox{mylistingbox}{
  colback=white,
  colframe=white,
  boxsep=0pt,
  boxrule=0pt,
  left=0pt,
  right=0pt,
  top=0pt,
  bottom=0pt,
  arc=0pt,
  outer arc=0pt,
  listing only
}

\begin{document}

\conference{IIR2023: 13th Italian Information Retrieval Workshop, 8th - 9th June 2023, Pisa, Italy}

\copyrightyear{2023}
\copyrightclause{Copyright for this paper by its authors. Use permitted under Creative Commons License Attribution 4.0 International (CC BY 4.0).}

\conference{IIR2023: 13th Italian Information Retrieval Workshop, June 8th - 9th, 2023, Pisa, Italy}

\title{Fauno: The Italian Large Language Model that will leave you senza parole!}


\author[1]{Andrea Bacciu}[%
email=bacciu@diag.uniroma1.it,
url=https://andreabac3.github.io/
]

\author[1]{Giovanni Trappolini}[%
]
\author[2]{Andrea Santilli}[%
]
\author[2]{Emanuele Rodolà}[%
]
\author[1]{Fabrizio Silvestri}[%
]



\address[1]{Sapienza University - Department of Computer, Control and Management Engineering}
\address[2]{Sapienza University - Computer Science Department}


\begin{abstract}
This paper presents Fauno, the first and largest open-source Italian conversational Large Language Model (LLM). 
Our goal with Fauno is to democratize the study of LLMs in Italian, demonstrating that obtaining a fine-tuned conversational bot with a single GPU is possible.
In addition, we release a collection of datasets for conversational AI in Italian.
The datasets on which we fine-tuned Fauno include various topics such as general question answering, computer science, and medical questions.
We release our code and datasets on \url{https://github.com/RSTLess-research/Fauno-Italian-LLM}
\end{abstract}


\maketitle

\section{Introduction}

Recently, there has been significant progress in enhancing the capabilities of natural language processing (NLP) methods. Notably, ChatGPT \citep{chatgpt} and GPT-4 \cite{openai2023gpt4} have emerged as highly proficient chat models. These models exhibit an impressive aptitude for comprehending and producing human-like responses across various domains. Consequently, they have gained popularity in customer support, virtual assistants, and social media moderation applications. However, these models are often only accessible through restricted APIs, which hampers new research and development. Additionally, the limited availability of chat models creates challenges for researchers and practitioners, impeding the growth of the NLP community. This problem is further aggravated by the scarcity of publicly accessible and high-quality chat datasets for analyzing and improving these models.
Recently, Baize \cite{xu2023baize}, whose depiction can be seen in Figure \ref{fig:baize-self-chatting}, has been introduced to tackle these challenges effectively, utilizing the capabilities of ChatGPT to generate a top-notch chat dataset comprising multiple conversational turns automatically. Their approach involves employing ChatGPT to engage in a simulated dialogue with itself, mimicking both user and AI responses. This generated dataset is a valuable resource for training and evaluating chat models in the context of multi-turn conversations. Additionally, by specifying an initial dataset, they extract samples from specific domains and fine-tune chat models to specialize in technology or healthcare. To fine-tune Large Language Models (LLM) with limited resources, they adopt an efficient tuning technique that maximizes the utilization of available computational capabilities. This strategy allows state-of-the-art language models to adapt and perform well in resource-constrained scenarios. Their primary objective is to enhance an open-source LLM, LLaMA \cite{touvron2023llama}, which they believe has the potential to serve as an accessible alternative to proprietary chat models. By fine-tuning LLaMA with their generated chat dataset, they introduce a new model called Baize (Bái zé), inspired by a legendary creature from Chinese folklore known for its linguistic prowess and vast knowledge. 
Baize reaches incredible performance; however, as with many other models in this category, its performance heavily depends on the language considered \cite{ducel2022we}.
In particular, these models tend to prefer the English language since this is the language they have seen the most during their training.
Their performance in other languages, such as Italian, is vastly inferior, as it can be easily checked empirically by merely using these tools.
For this reason, inspired by Latin mythology, in this paper, we introduce Fauno, the first and the largest open-source Italian conversational LLM.
To build Fauno, we translated several datasets originally used by Baize in Italian. We use them to fine-tune and train our model, reaching good generation scores, as highlighted in later sections of this paper.
Our goal is twofold.
First, we want to provide the Italian (and the international) research community with an LLM that is highly proficient in Italian. We hope this will enhance and enable new research on this particular language by surpassing the current state-of-the-art capabilities of Italian.
Second, we want to provide practitioners with a valid conversational AI able to deal with the Italian language.
We provide an example for this latter case to hint at this tool's potential.
Italy has one of the largest populations of elderly people.
In the last twenty years, the number of people older than eighty has doubled, according to the Italian statistical entity (ISTAT).
Furthermore, roughly 1.2 million people are older than 65 and live alone.
These people are known in the medical literature to suffer from loneliness, which aggravates their health issues.
With tools like Fauno, however, one could offer a conversational AI as a telehealth service.
While not being a substitute for a human, this could improve the quality of life of millions of people and reduce the weight of the national health system.
This is just one example of applications, and we hope our model can spark many such projects and research.
Fauno can be operational on accessible hardware, thus expanding accessibility to a broader range of practitioners and researchers.

\section{Related Work}
\subsection{Language Models for Chat}
Transformers \cite{vaswani2017attention}, introduced initially as a machine translation system, have had an arguably unprecedented impact on the AI world.
Transformers and deep learning models, in general, are now used not only for NLP \cite{devlin2018bert,raffel2020exploring,trappolini2023mmndb,conia2021unifying}, but also other fields such as image processing \cite{dosovitskiy2020image}, audio \cite{dhariwal2020jukebox,agostinelli2023musiclm,barnabo2023cycledrums}, 3D data \cite{zhao2021point,trappolini2021shape,halimi2020towards}, recommendation systems \citep{kang2018self, bacciu2023integrating}.
Language Models (LMs), based on the transformer architecture, have gained significant attention in Natural Language Processing due to their ability to produce coherent human-like text. \citet{radford2019language} showed that LMs could learn a variety of tasks without direct human supervision but using self-supervision on large, noisy web corpora. Several works have developed this idea, further scaling the data and model size \cite{brown2020language, bommasani2021opportunities, hoffmann2022training}. However, all these models are not explicitly designed with a conversational objective or training curriculum. \citet{zhang2019dialogpt} introduced DialoGPT, which is a fine-tuned version of GPT-2 on Reddit data; \citet{adiwardana2020towards} proposed Meena, a 2.6B model trained with conversational data from social media; \citep{thoppilan2022lamda} proposed LaMDA, a 137B model trained on public dialog data. Recently OpenAI released ChatGPT \citep{chatgpt}, a model optimized for conversational interaction trained with Reinforcement Learning with Human Feedback \cite{ouyang2022training}. With GPT-4~\citep{openai2023gpt4}, this approach was further improved with additional reasoning capabilities. Recently, \citet{xu2023baize} introduced Baize as a finetuned version of LLaMA \cite{touvron2023llama} for conversation by training the model on dialogs generated by ChatGPT. Our approach stems from this work and extends it to the Italian language.

\subsection{Language Models for Italian}
To the best of our knowledge, no LMs are trained explicitly with a conversational objective or training curriculum for the Italian language. The currently available LMs for Italian include GePpeTto\cite{DBLP:conf/clic-it/MatteiCDNG20}, a GPT-2 model finetuned on Italian Wikipedia and the ItWac corpus \cite{baroni2009wacky}; IT5 \cite{sarti2022it5}, an Italian version of T5 trained on a cleaned version of the Italian mC4 corpus\cite{xue-etal-2021-mt5}; BART-IT \cite{fi15010015}, an Italian version of BART\cite{lewis-etal-2020-bart} trained on the same data mixture of IT5; Camoscio \cite{camoscio}, an Italian version of Stanford Alpaca \citep{alpaca} trained on the Alpaca dataset translated to Italian.

\subsection{Parameter-Efficient Tuning}
Typical fine-tuning involves training all parameters of an LLM, which can be inefficient as the number of parameters increases. Various techniques have been proposed to address this issue under the umbrella term of Parameter-Efficient Fine-Tuning (PEFT). These techniques include Adapters\cite{adapter}, which adds a trainable layer inside the Transformer while freezing the others; BitFit\cite{bitfit}, which only tunes bias terms in the linear layers; Prompting approaches like Prefix Tuning \cite{prefixtuning}, P-Tuning \cite{liu-etal-2022-p} or Prompt Tuning \cite{lester2021power} which fine-tune the optimal prompt parameters for a task; or LoRA \cite{lora} which trains low-rank matrices while keeping all the remaining parameters frozen. This is a very active research area, and we refer the reader to \citet{Zhuang2023ASO} for an up-to-date overview.
Recently the community leveraged these techniques to train low-cost versions of Stanford Alpaca \citep{alpaca} under the name of Alpaca-LoRA\footnote{\url{https://github.com/tloen/alpaca-lora}}.

\section{Conversational AI Datasets in Italian}
In this section, we report the data we use to fine-tune Fauno and the data generation and translation process.
We train Fauno on extensive Italian synthetic datasets, encompassing a wide range of fields such as medical data (MedQuad~\cite{medquad}), technical content from StackOverflow, Quora discussions, and Alpaca~\citep{alpaca} data translated into Italian with 47.000, 57.000, 54.000 and, ~22.000 dialogues, respectively.
The synthetic generation of the data is described in Section \ref{sec:data-generation}.

\begin{wrapfigure}{l}{0.4\textwidth}
    \centering
    \includegraphics[width=0.4\textwidth]{{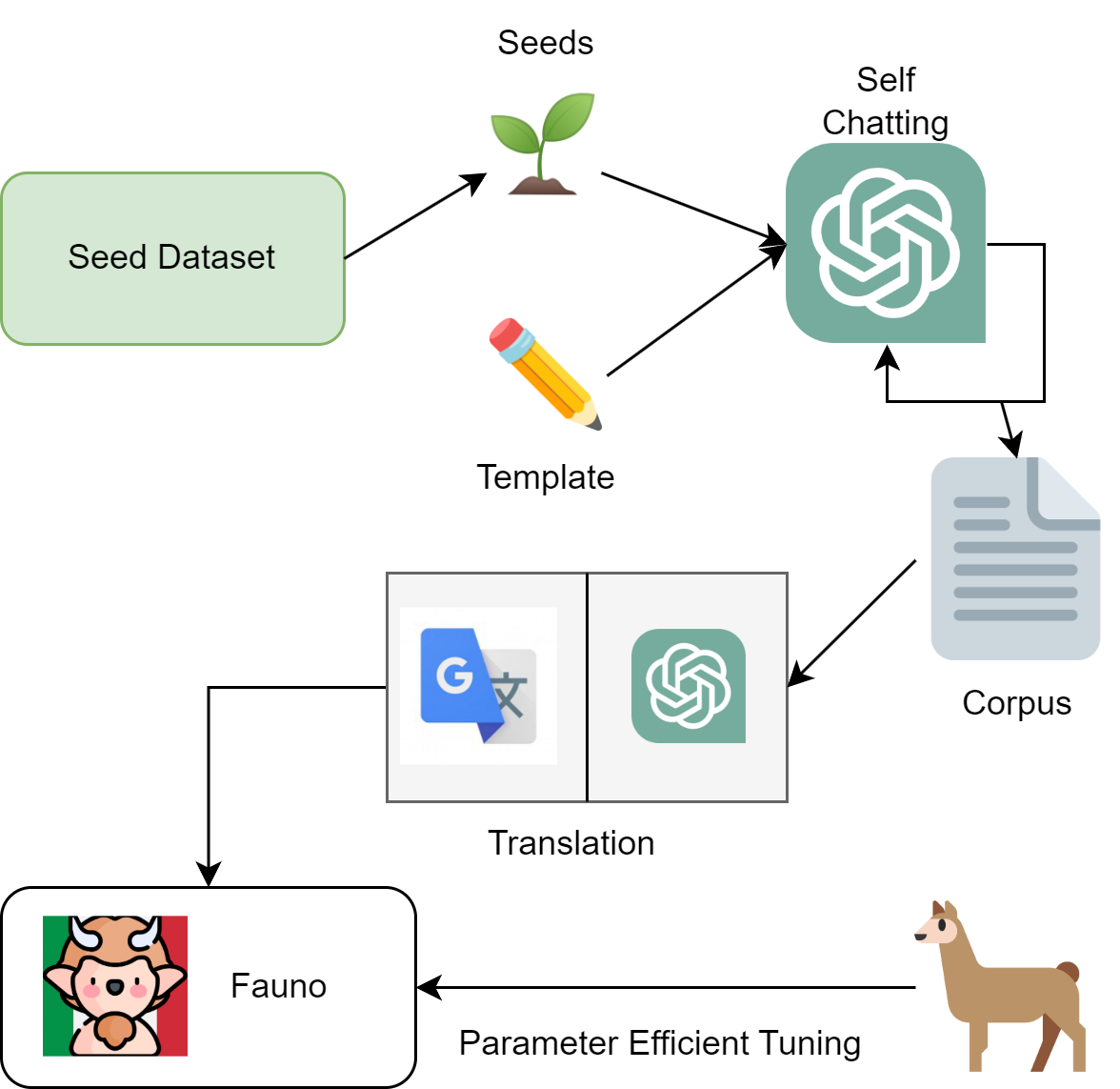}}
    \label{fig:baize-self-chatting}
\end{wrapfigure}

\subsection{Data Generation} \label{sec:data-generation}
To produce conversational AI Datasets, we inherit Xu et al. 2023~\cite{xu2023baize}'s dataset generation methodology.
This methodology consists of allowing ChatGPT to engage in self-chatting to produce a conversation between a user and a helpful AI assistant in a conversational format.
This self-chatting process is initiated with a question (or seed) from real users. For example, \cite{xu2023baize} takes as seed questions froms Quora\footnote{\url{https://huggingface.co/datasets/quora}} and Stack Overflow\footnote{\url{https://huggingface.co/datasets/pacovaldez/stackoverflow-questions}}.
The generation methodology is shown in Figure \ref{fig:baize-self-chatting}.

\subsection{Translation}
To translate Fauno's data, we use two different translation systems to maximize translation accuracy and quality. 
Jiao et al. (2023) \cite{jiao2023chatgpt} have demonstrated that Google Translate is state-of-the-art among commercial translation systems. Nevertheless, Google Translate encounters significant issues when handling text incorporating natural language utterances and programming code. For instance, when translating a code snippet like "for i in range(10)" into Italian, Google Translate would incorrectly render it as "per i in intervallo(10)", which disrupts the syntax and functionality of the original code. This is due to Google Translate's inability to discern between natural language and programming code. To address this limitation, we opted to utilize ChatGPT 3.5 to translate datasets containing programming code, as it is better suited for preserving the integrity of the code's syntax and structure.
Hence, we translate Alpaca and StackOverflow with ChatGPT 3.5 and the remaining datasets with Google Translate.
We release the translated dataset of StackOverflow\footnote{\url{https://huggingface.co/datasets/andreabac3/StackOverflow-Italian-Fauno-Baize}}, MedQuad\footnote{\url{https://huggingface.co/datasets/andreabac3/MedQuaAD-Italian-Fauno-Baize}}, Quora\footnote{\url{https://huggingface.co/datasets/andreabac3/Quora-Italian-Fauno-Baize}} and Alpaca\footnote{\url{https://huggingface.co/datasets/teelinsan/camoscio_cleaned}} in the Huggingface hub.
The cost of translating the StackOverflow dataset using the ChatGPT-3.5 API is ~\$80, while the Alpaca Dataset translation costs around \$30. For other datasets, we utilize Google's free translation API\footnote{\url{https://pypi.org/project/googletrans/}}.

\section{Train setup}
We use the same hyper-parameters of \cite{xu2023baize} with a single epoch and batch size of 128 examples. 
We perform our experiments on a workstation equipped with an Intel Core i9-10940X (14-core CPU running at 3.3GHz) and 256GB of RAM, and a single Nvidia RTX A6000 with 48GB of VRAM.
The training time for the 7B and 13B models are 19 and 35 hours, respectively.

\section{Evaluation}
As done by Xu et al. 2023~\cite{xu2023baize} and Camoscio \cite{camoscio}, we rely on a qualitative analysis of Fauno's outputs.
In Table \ref{tab:nlp_comparison}, we compare ChatGPT 3.5, Camoscio, and Fauno.


\begin{table}[h]
\centering
\scalebox{0.85}{
\begin{tabularx}{\textwidth}{@{}lX@{}}
\toprule
\textbf{Model} & \textbf{Answer} \\ \midrule
\textbf{Question} & \textbf{Qual è il senso della vita? }\\
\midrule
ChatGPT & La domanda sul senso della vita è stata oggetto di riflessione filosofica e teologica per secoli, e non esiste una risposta definitiva o concorde. Tuttavia, molti individui trovano significato e scopo nella vita attraverso diverse vie. Alcuni cercano il senso nella connessione umana e nelle relazioni significative, altri nella realizzazione personale e nello sviluppo delle proprie passioni. Altri ancora si rivolgono alla spiritualità o alla fede per trovare un senso più profondo. In generale, il senso della vita può essere un'esperienza altamente individuale e soggettiva, e ognuno potrebbe trovare una risposta diversa a questa domanda.\\
\midrule
Camoscio & Il senso della vita è trovare la felicità e la soddisfazione nella vita.\\
\midrule
Fauno & \small{Il senso della vita è una domanda popolare e complessa che ha suscitato dibattito in molte culture e generazioni. Alla fine, la comprensione di qual è il senso della vita dipende dalla prospettiva e dalle convinzioni personali di ogni singolo individuo. Alcune persone possono vedere il senso della vita in termini di un significato unico e globale, mentre altri possono sforzarsi di risolvere il mistero della vita trovando autentico benessere e piacere nei loro giorni quotidiani.}\\
\midrule
\textbf{Question} & \textbf{Come posso copiare un file da un server tramite ssh? }\\
\midrule
ChatGPT & Per copiare un file da un server tramite SSH, puoi utilizzare il comando "scp" (Secure Copy). Ecco come puoi farlo:
Apri una finestra di terminale sul tuo computer locale.
Assicurati di avere il client SSH installato sul tuo sistema. Se stai utilizzando un sistema operativo basato su Unix (ad esempio Linux o macOS), lo avrai probabilmente già installato. Se stai utilizzando Windows, puoi utilizzare un client SSH come PuTTY o installare un'implementazione di SSH come OpenSSH.
Utilizza il seguente comando "scp" per copiare il file dal server al tuo computer locale:
scp username@indirizzo\_server:percorso\_del\_file\_locale percorso\_di\_destinazion\\
\midrule
Camoscio & Per copiare un file da un server tramite ssh, è necessario utilizzare l'opzione "scp" dell'utente ssh. Per esempio, per copiare un file dal server "server.example.com" all'utente "utente.example.com", l'utente dovrebbe utilizzare l'opzione "scp server.example.com:utente.example.com".\\
\midrule
Fauno & Per copiare un file da un server tramite ssh, puoi utilizzare il seguente comando: 

scp username@ip\_dello\_server:percorso/della/sorgente/del/file /percorso/della/destinazione/del/file 

Sostituisci i dettagli con il tuo nome utente, il nome dell'host, il percorso della sorgente e il percorso finale del file e clicca su "Avanti". Il file verrà copiato dal server alla tua macchina. C'è qualcos'altro di cui hai bisogno?\\

\bottomrule
\end{tabularx}
}
\caption{Qualitative analysis of ChatGPT-3.5, Camoscio, and Fauno 7B}
\label{tab:nlp_comparison}
\end{table}



\section{Conclusion}
This work shows that it is possible to democratize the research for LLMs in multiple languages with limited resources.
Furthermore, we believe that the release of the Italian datasets for conversational AI will help researchers to develop further Large Language Models.

\clearpage
\newpage

\section*{ACK}
This work was partially supported by projects FAIR (PE0000013) and SERICS (PE00000014) under the MUR National Recovery and Resilience Plan funded by the European Union - NextGenerationEU and by ERC Starting Grant No. 802554 (SPECGEO) and PRIN 2020 project n.2020TA3K9N "LEGO.AI". Supported by the ERC Advanced Grant 788893 AMDROMA,  EC H2020RIA project “SoBigData++” (871042), PNRR MUR project  IR0000013-SoBigData.it. 

\bibliography{sample-ceur}

\end{document}